\documentclass[10pt,twocolumn,letterpaper]{article}
\usepackage{ijcb}
\usepackage{times}
\usepackage{epsfig}
\usepackage{graphicx}
\usepackage{amsmath}
\usepackage{amssymb}
\usepackage{subcaption}
\usepackage{tabularx}
\usepackage{threeparttable}
\usepackage{multirow}
\usepackage{multicol}
\usepackage{comment}
\usepackage{makecell}
\usepackage{enumitem}
\usepackage{enumerate}
\usepackage{MnSymbol}
\usepackage{url}
\usepackage{booktabs}

\newcolumntype{L}{>{\raggedright\arraybackslash}X}
\ijcbfinalcopy % *** Uncomment this line for the final submission

\begin{document}

%%%%%%%%% TITLE
\title{Iris Liveness Detection Competition (LivDet-Iris) -- The 2023 Edition}

\author{Patrick Tinsley$^{*1\dag}$, Sandip Purnapatra$^{*2\dag}$, Mahsa Mitcheff$^{1\dag}$, Aidan Boyd$^{1\dag}$, Colton Crum$^{1\dag}$
\and
Kevin Bowyer$^{1\dag}$, Patrick Flynn$^{1\dag}$, Stephanie Schuckers$^{2\dag}$, Adam Czajka$^{1\dag}$, Meiling Fang$^{3\ddag}$
\and
Naser Damer$^{3\ddag}$, Xingyu Liu$^{4\ddag}$, Caiyong Wang$^{4\ddag}$, Xianyun Sun$^{4\ddag}$, Zhaohua Chang$^{4\ddag}$, Xinyue Li$^{4\ddag}$
\and
Guangzhe Zhao$^{4\ddag}$, Juan Tapia$^{5\ddag}$, Christoph Busch$^{5\ddag}$, Carlos Aravena$^{6\ddag}$, Daniel Schulz$^{6\ddag}$
\and
$^1$University of Notre Dame, IN, USA; $^2$Clarkson University, NY, USA;
$^3$Fraunhofer Institute \and for Computer Graphics Research IGD, Germany; $^4$School of Electrical \and and Information Eng., Beijing University of Civil Engineering and Architecture (BUCEA), China
\and
$^5$Hochschule Darmstadt (HDA), Germany; $^6$IDVisionCenter, Germany
\and
\small{$^*$Equal contribution;~$^\dag$Organizers;~$^\ddag$Competitors; Corresponding emails: {\tt ptinsley@nd.edu}, {\tt purnaps@clarkson.edu}}
}

\maketitle

\pagestyle{empty}
%%%%%%%%% ABSTRACT
\begin{abstract}
\vskip-3mm
% (first attempt)
This paper describes the results of the 2023 edition of the ``LivDet'' series of iris presentation attack detection (PAD) competitions.
New elements in this fifth competition 
% of the series 
include (1) GAN-generated iris images as a category of presentation attack instruments (PAI), and (2)  an evaluation of human accuracy at detecting PAI as a reference benchmark.
Clarkson University and the University of Notre Dame contributed image datasets for the competition, composed of samples representing seven different PAI categories, as well as baseline PAD algorithms.
Fraunhofer IGD, Beijing University of Civil Engineering and Architecture, and Hochschule Darmstadt contributed results for a total of eight PAD algorithms to the competition.
%, representing a variety of interesting technical approaches.
Accuracy results are analyzed by different PAI types, and compared to human accuracy. 
Overall, the Fraunhofer IGD algorithm, using an attention-based pixel-wise binary supervision network, showed the best-weighted accuracy results (average classification error rate of 37.31\%), while the Beijing University of Civil Engineering and Architecture's algorithm won when equal weights for each PAI were given (average classification rate of 22.15\%). These results suggest that iris PAD is still a challenging problem.
\end{abstract}

\vskip-5mm

%%%%%%%%% BODY TEXT
\section{Introduction}

LivDet-Iris 2023 is the fifth competition in the LivDet-Iris series offering (a) an independent assessment of the current state of the art in iris PAD \cite{czajka2018presentation,boyd2020iris} algorithms, and (b) an evaluation protocol, including datasets of bona fide and spoofed iris images, including the most recent PAI.
% that can be followed by researchers 
This means that, after the competition is closed, researchers can still compare their solutions with LivDet-Iris winners and baselines. LivDet-Iris 2023 has been included in the official IJCB 2023 competition list\footnote{\scriptsize\url{https://ijcb2023.ieee-biometrics.org/competitions/}}.

Each edition of LivDet-Iris introduces novel elements to the competition.
%, as it's also the case with LivDet-Iris 2023. For the first time,
For the 2023 edition, we have included synthetic iris images generated by Generative Adversarial Networks (StyleGAN2-ADA and StyleGAN3), trained for various numbers of iterations, translating to various levels of realism of the generated samples. Also, for the first time in the LivDet-Iris series, we asked human subjects to classify all test images, and compared humans' accuracy with that of the baseline and submitted algorithms.
%machine's (submissions + baselines) accuracy.

LivDet-Iris 2023 proposed three types of participation (and evaluation), organized into three parts:
\begin{itemize}[noitemsep]
\item {\bf Part 1 (Algorithms-Self-Tested)} involved the self-evaluation (that is, done by competitors on the sequestered and never-published-before test data) of 13,332 ISO-compliant bona fide and PAI samples. 
\item {\bf Part 2 (Algorithms-Independently-Tested)} involved the evaluation of the software solutions submitted by the competitors and performed by the organizers. % on a sequestered dataset similar in terms of attack types to the one used in Part 1. 
\item {\bf Part 3 (Systems)} involved the systematic testing of submitted iris recognition systems based on physical artifacts presented to the sensors.
\end{itemize}

Competitors could participate in one, two or all three parts. This edition of LivDet-Iris obtained eight submissions from three research teams for Part 1. Parts 2 and 3 did not receive submissions. %Due to 
Since we chose two different ways of calculating the Attack Presentation Classification Error Rate (APCER), either as a weighted average or non-weighted average (where weights are proportional to the number of samples), we decided to announce two winners, depending on the calculation method. The
%BUCEA
Beijing University of Civil Engineering and Architecture team wins when APCER is not weighted by the number of samples with the Average Classification Error Rate (ACER) equal to 22.15\%. When weighted APCER is calculated, the Fraunhofer IGD team wins with ACER = 37.31\%. It is important to note that ACER levels from both winning teams are relatively large, and comparable with the results obtained for baseline models, which were trained with different PAI than the LivDet-Iris 2023 test set. The immediate conclusion here is that open-set iris PAD remains challenging \cite{BoydIrisPAD}.

The {\bf main contributions} of the LivDet-Iris 2023 competition can be summarized as follows:
%
%\begin{itemize}[noitemsep]
%\item 
(1) {\bf Independent assessment} of eight iris PAD algorithms in a competition setup; (2) The {\bf test dataset} composed of 13,332 samples (with PAI labels and all baseline scores), representing eight PAI types (printouts, cosmetic contact lenses, printouts with contact lens on top, eye dome on a printout, image displayed on e-ink device, doll eyes with and without contact lenses on them, and synthetically-generated images with StyleGAN2 and StyleGAN3); (3) {\bf Human examination} results on the test dataset.
%\end{itemize}

\section{Previous LivDet-Iris Competitions}

The idea of running an independent evaluation of biometric PAD algorithms was pioneered by the {\it Fingerprint Liveness Detection Competition} in 2009 \cite{Marcialis_ICIAP_2009}. This was extended in 2013 to the iris recognition modality with the first edition of LivDet-Iris \cite{Yambay_BTAS_2013}. The next three editions were held in 2015 \cite{Yambay_ISBA_2017}, 2017 \cite{Yambay_IJCB_2017} and 2020 \cite{das2020iris}, with LivDet-Iris 2023 being the fifth competition in the LivDet-Iris series. 

Each edition offered novel elements compared to previous editions. While the 2013 edition was essentially organized in a closed set scenario (properties of the test samples were known), the 2015 edition held out some contact lens brands for testing only, making them unknown to the participants. The 2017 edition expanded the number of sensors used to collect the data, the number of teams preparing the data (four, compared to three in the 2015 edition), and the samples. The 2020 edition introduced PAI types never used in previous competitions: post-mortem iris samples, irises presented on an electronic display, and prosthetic and printed eyes with additional contact lenses on top of these artifacts. It also included a few baseline algorithms, and the entire competition was organized in an open-set scenario. In total, thirteen teams have competed in the four previous LivDet-Iris competitions.

The main conclusion from each competition was that the iris PAD was far from being considered a solved problem. While the biometric community was able to achieve almost perfect PAD performance with older LivDet-Iris test benchmarks, new test sets used in the consecutive editions allowed for average classification error rates from approx. 3.6\% (2017) to approx. 29\% (2020). Results obtained this year only confirm that the iris PAD remains challenging when the exact properties of PAI samples (which are compliant with ISO/IEC 19794-6) are unknown to the participants.

\section{Experimental Setup}

\subsection{Submission Protocol}

Participants of Part 1 were offered the self-evaluation package with the data (after executing two data-sharing license agreements and sending them to the organizers). The self-evaluation package included a small ``instructional'' subset consisting of 60 images with correct labels (PAI/bona fide). The goal of this subset was to present the format and nature of the test data. The package included also the ``test'' data, without ground truth labels, to be used in actual self-evaluation, and came with instructions on how to prepare the CSV files with PAD scores, and how to submit them to the organizers. The participants were not allowed to train their algorithms, or otherwise use the test subset in the algorithm's design. The generated iris PAD scores should be in the range of [0,100] for all of the test samples, where 100 is the maximum degree of liveness, and 0 means that the image is a PAI. If the image cannot be processed, the participants were asked to mark it with a score of -1000.

Part 2 assumed submission executables to the organizers, and Part 3 required to submit the entire systems for testing. There were no Part 2 and Part 3 submissions.

% Sample images 6,5
\subsection{Competition Datasets} 

% Adam's version:
\begin{figure*}[!ht]
\centering
\subcaptionbox{\centering Bona fide Iris}
{\includegraphics[width=0.16\textwidth]{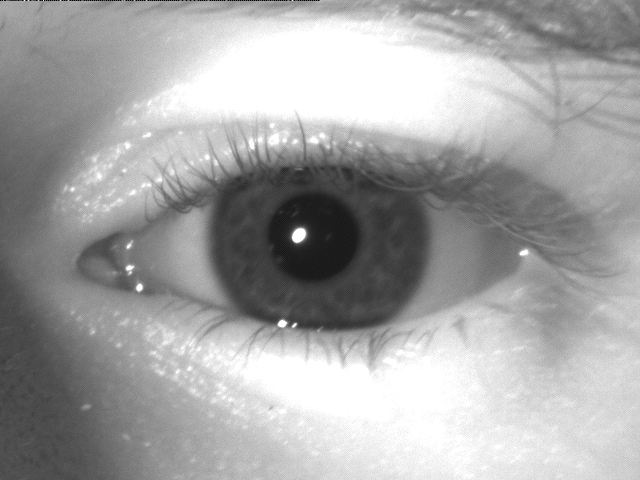}}
\hfill
\subcaptionbox{\centering Paper printout}
{\includegraphics[width=0.16\textwidth]{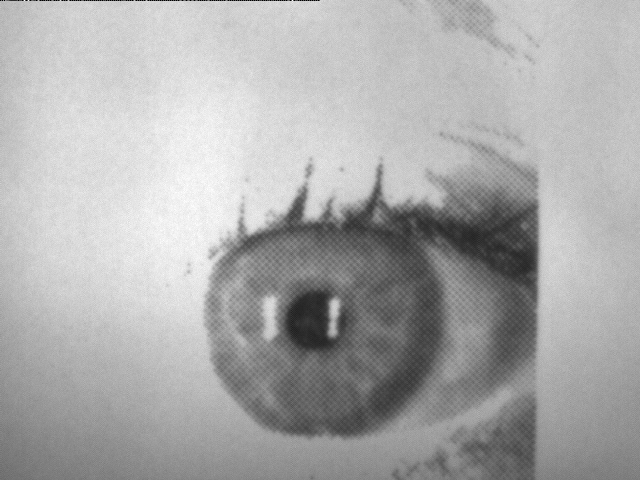}}
\hfill
\subcaptionbox{\centering Cosmetic contact lens on bona fide eye}
{\includegraphics[width=0.16\textwidth]{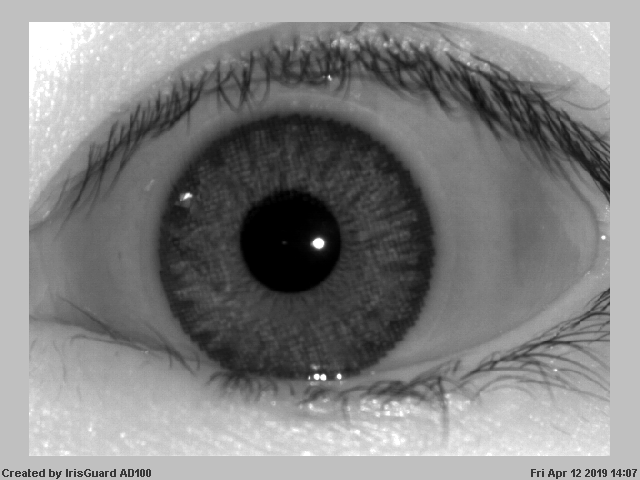}}
\hfill 
\subcaptionbox{\centering Cosmetic contact lens on printed eye}
{\includegraphics[width=0.16\textwidth]{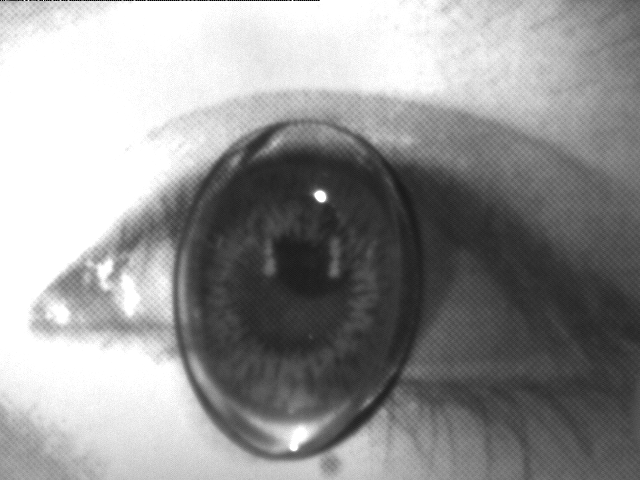}}
\hfill
\subcaptionbox{\centering Eye dome on the printed eye}
{\includegraphics[width=0.16\textwidth]{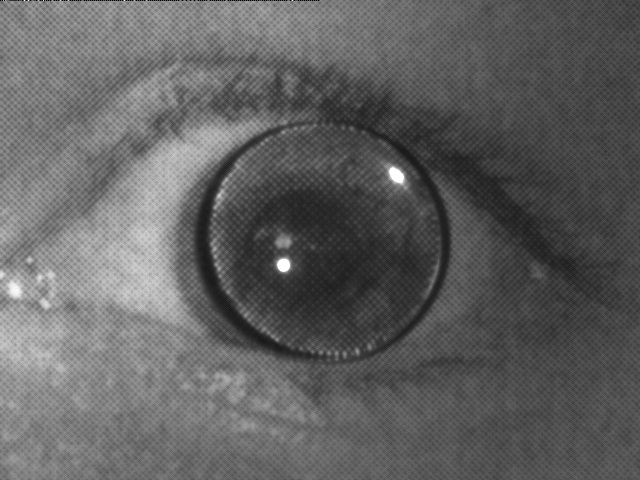}}
\hfill
\subcaptionbox{\centering Doll eye}
{\includegraphics[width=0.16\textwidth]{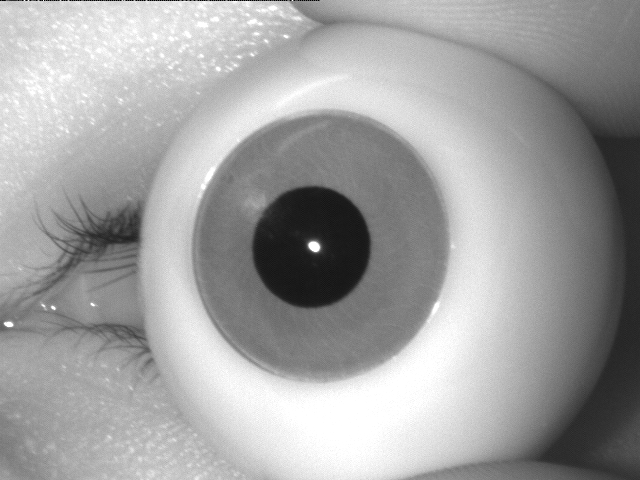}}\newline
\hfill\subcaptionbox{\centering Electronic display}
{\includegraphics[width=0.16\textwidth]{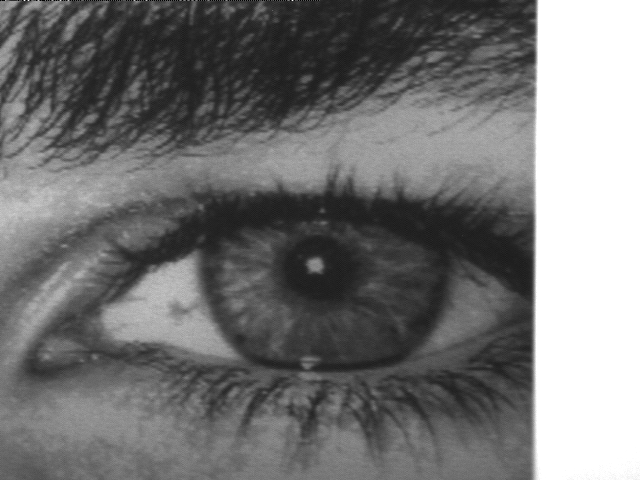}}
\hskip0.5mm
\subcaptionbox{\centering Cosmetic contact lens on the doll eye}
{\includegraphics[width=0.16\textwidth]{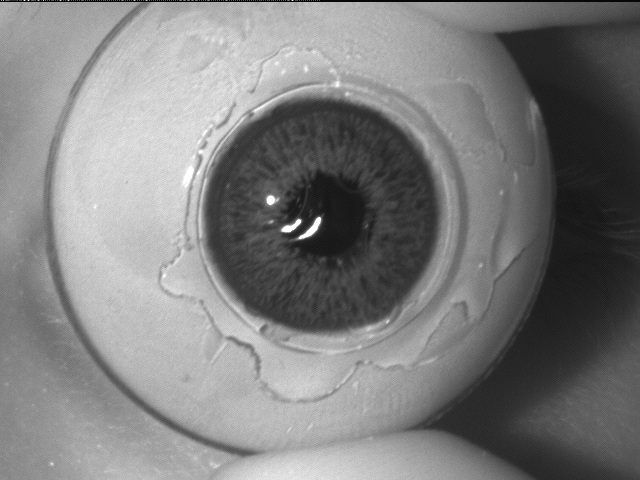}}
\hskip0.5mm
\subcaptionbox{\centering Synthetic Iris \newline(Low Quality)}
{\includegraphics[width=0.16\textwidth]{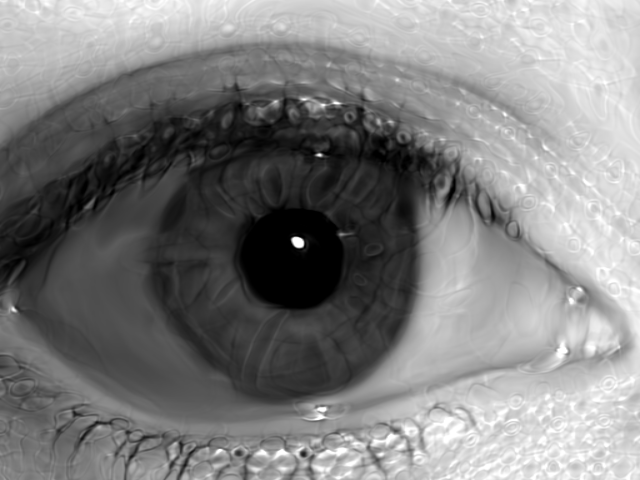}}
\hskip0.5mm
\subcaptionbox{\centering Synthetic Iris \newline(Medium Quality)}
{\includegraphics[width=0.16\textwidth]{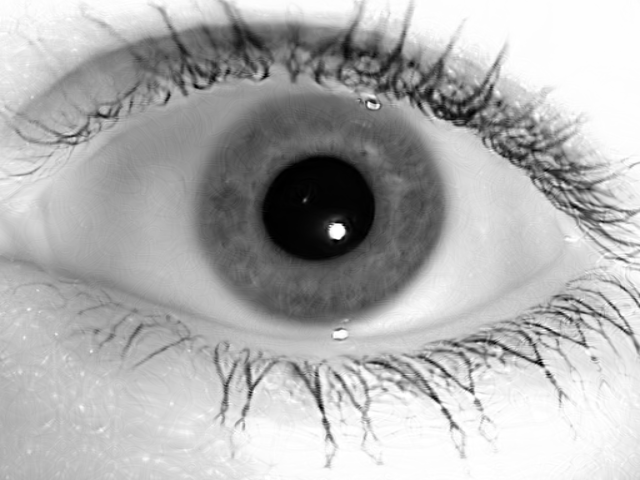}}
\hskip0.5mm
\subcaptionbox{\centering Synthetic Iris \newline(High Quality)}
{\includegraphics[width=0.16\textwidth]{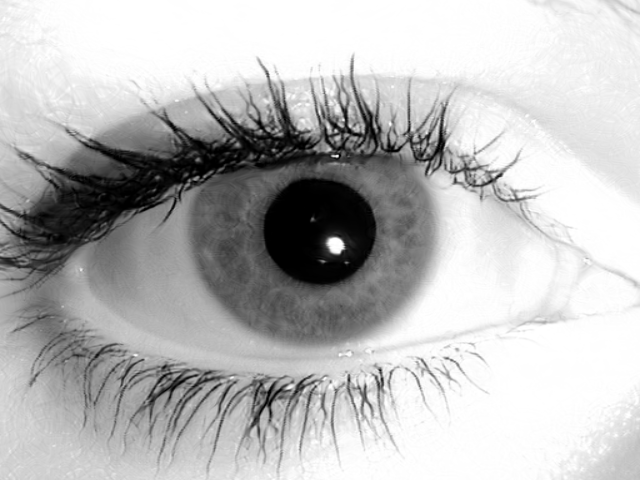}}
\hfill
\caption{Example images of a bona fide iris image (a) and all presentation attack instruments (PAI) used in the test set (b-k).}
\label{fig:Example_images}
\end{figure*}

\subsubsection{Test Data}

The test data\footnote{A copy of the test benchmark can be requested at \url{https://cvrl.nd.edu/projects/data/#livdet-iris-2023-part1}} consisted of 13,332 iris images (6,500 bona fide, 6,832 PAI samples), as summarized in Table~\ref{table:Dataset}. Example test images per PAI category can be seen in Figure ~\ref{fig:Example_images}. The following eight PAI categories were included:

\begin{itemize}[leftmargin=*, noitemsep]
\item{\bf Printed eyes (PE)} 522 samples were collected using five different printers (Epson Stylus Pro 9900, HP M652, Xerox C60, OKI MB-471, Cannon Super G3) and two different printing qualities (``office'' and ``professional''). Additionally, two different printing papers were used for this PAI: matte and glossy. All images in this PAI category were captured with an Iris ID iCAM7000 sensor.

\item{\bf Textured contact lenses (CL)} 4 samples were acquired using LG IrisAccess 4000 and IrisGuard AD100 sensors under varying illumination setups.

\item{\bf Eyes displayed on Kindle e-Ink (ED)} 40 near-infrared (NIR) samples from this PAI category were captured by the Iris ID iCAM7000 sensor.  

\item{\bf Fake/Prosthetic/Printed Eyes with Add-Ons (FP)} This PAI category has five sub-categories of PAIs for a total of 266 samples, captured in NIR by the Iris ID iCAM7000 sensor, with a non-uniform distribution of samples in each category: (i) \textbf{Textured Contacts on Printed Eyes:} patterned contact lenses added on top of the printed eye images; (ii) \textbf{Textured Contacts on Doll Eyes:} patterned contact lens put on the iris area of plastic doll eyes; (iii) \textbf{Clear Contacts on Printed Eyes:} transparent contact lens added on top of the printed eye images; (iv) \textbf{Eye Dome on Printed Eyes:} transparent 3D plastic eye domes added on top of the printed eye images; (v) \textbf{Doll Eyes:} Fake eyes of two different types -- Van Dyke Eyes (higher iris quality details) and Scary Eyes (plastic fake eyes with a simple pattern on the iris region). Different color variations of both types of fake eyes were included.
\end{itemize}

% \end{itemize}

\paragraph{Synthetic LQ/MQ/HQ}
%This year's edition of the iris liveness detection competition
LivDet-Iris 2023 also included GAN-generated synthetic images.
% \footnote{Although ``sensor-less'' synthetic samples do not explicitly qualify as presentation attacks, PA and liveness detection methods can still be applied to and benefit from computer-generated images.}.
\footnote{Synthetic samples are more likely to be used as part of an injection attack or other form of digital attack, rather than as presentation attacks.  However, it has been show that PA and liveness detection methods may also be useful for detecting computer-generated, synthetic images.}
Two state-of-the-art generative models (StyleGAN2-ADA and StyleGAN3~\cite{karras2020sg2ada, karras2021sg3}) were trained on a custom dataset containing 407,425 bona fide samples. Before initiating the training process, sensor-based artifacts in the images were removed (for instance, in the case of IrisGuard AD100 sensors, labels in the bottom-left image corner reading ``Created by IrisGuard, Inc.'') by adding gray bars spanning the width of the image. Additionally, since StyleGAN requires square images for training, images were resized from their native resolution of $640 \times 480$ to $512 \times 512$. After the training process was complete, 1000 images were generated for each StyleGAN model at three different training ``snapshots'', for a total of 6000 synthetic images. The three snapshots correspond to low, medium, and high-quality samples, labeled \textbf{LQ}, \textbf{MQ}, and \textbf{HQ}, respectively. Generators from earlier snapshots in the training process tended to produce samples with ``bubble'' artifacts and were thus dubbed LQ since these artifacts might more easily be learned by a PAD model.
%Contrarily,
By contrast, MQ/HQ samples generated by models at later snapshots are virtually indistinguishable from bona fide samples, featuring detailed iris textures and believable eyelashes, eyebrows, and textures.

One problem with the generated data was the presence of gray bars around images captured by IrisGuard and Iris ID sensors (as a consequence of iris repositioning done by the sensors' algorithms to place the pupil in the center). These gray bars were very well, but not perfectly, imitated by the StyleGAN models. We thus replaced the gray bars generated by StyleGANs with ideal gray bars (as they would originate from the sensors) of intensity equal to 192. This post-generation task did not interfere with iris texture or ocular information but made the synthetic images more similar to bona fide images.

\subsubsection{Instructional and Train Data}

In addition to the unlabeled test data, the competitors were also given 60 labeled iris samples (30 bona fide, 30 PAIs). This small sample was created solely to demonstrate the format and nature of the test data. As in the past (2020) edition of LivDet-Iris, there was no official training dataset. Instead, the participants were free to use any data they considered useful to make their submissions stronger. This creates a more realistic scenario, in which the design of effective PAD algorithms is not restricted by any specific training sample.

\begin{table*}[!ht]
\footnotesize 
\centering
\caption{LivDet - 2023 Iris Test Dataset Summary}
\label{table:Dataset}
\begin{tabular}{|c|c|c|c|}
\toprule
\textbf{Class} & \textbf{Presentation Attack Instruments (PAI)}  & \textbf{Sample Count} & \textbf{Sensor}  \\
\midrule
Bona fide & N/A & 6500 & LG 4000, AD 100, Iris ID iCAM7000 \\
% &  &  & Iris ID iCAM7000  \\
\hline
PAI & Printed Eyes (PE) & 522 & Iris ID iCAM7000 \\
\hline 
PAI & Textured Contact Lens (CL) & 4 & LG 4000, AD 100, Iris ID iCAM7000\\
% &  &  & Iris ID iCAM7000  \\
\hline
PAI & Electronic Display (ED) & 40 & Iris ID iCAM7000 \\
\hline 
PAI & Fake/Prosthetic/Printed Eyes with Add-Ons (FP) & 266 & Iris ID iCAM7000 \\
\hline
PAI & Synthetic Iris - Low Quality (LQ) & 2000 & N/A\\
\hline
PAI & Synthetic Iris - Medium Quality (MQ) & 2000 & N/A\\
\hline
PAI & Synthetic Iris - High Quality (HQ) & 2000 & N/A\\
\bottomrule

\end{tabular}
\end{table*}

\subsection{Performance Evaluation Metrics}
\label{evaluation}

% AC: we were using APCER and BPCER after defining them -- I am suggesting to move this section closer to the Introduction

LivDet-Iris 2023 follows the recommendations of ISO/IEC 30107-3:2017 \cite{ISO_IEC_301073:2017} in employing two basic PAD metrics in its evaluations: \textbf{Attack Presentation Classification Error Rate (APCER)}, the proportion of PAI presentations of the same PAI incorrectly classified as bona fide presentation, \ie PAI samples classified as bona fide, and \textbf{Bona fide Presentation Classification Error Rate (BPCER)}, the proportion of bona fide presentations classified as attack presentations, \ie bona fide samples classified as PAI.

Both the APCER and BPCER metrics are used to evaluate the algorithms. ISO/IEC 30107-3:2017 also recommends using the maximum value of APCER when multiple PAI categories are present in the case of system-level evaluation, which is primarily designed for industry applications. This, however, is inconsistent with our prior competitions \cite{yambay2014schuckers, 7947701, yambay2017livdet, das2020iris} and also our goal to consider the detection of all PAIs, and not to rank the competitors by looking at their worst-performing PAI. Thus, we introduced the sample-weighted average of APCER over all PAIs,
%\begin{itemize}[leftmargin=*, noitemsep]
%\item \textbf{Sample-Weighted Average of APCER} (APCER1), 
which is the average of APCER across all PAIs, weighted by the sample counts in each PAI category, reported as APCER1 in Table~\ref{table:Dataset}. Additionally, we calculated non-weighted APCER (reported as APCER2 in Table~\ref{table:Dataset}). Only for the {\bf purpose of competition ranking}, the Average Classification Error Rates (ACER1 and ACER2) were computed to select the winner(s) by averaging BPCER with either APCER1 or APCER2, respectively.  
% \item \textbf{Average Classification Error Rate (ACER1):} the average of APCER1 and BPCER.
%\end{itemize}
ACER was deprecated in ISO/IEC 30107-3:2017.

\subsection{Winner Selection}
% Explain two-winner scheme based on APCER weights

There were two ways of calculating the ACER (and thus the winners) this year. The first incorporated weighting the APCER by the relative number of samples per PAI type (as described above). The second weighted each PAI type equally, ignoring the fact that the test set was not balanced across PAI types. Since each of the ways of calculating the final ACER may have its justification, the organizers decided to announce two winners, depending on the ACER calculation procedure; these different ACERs are labeled ACER1 and ACER2, respectively.

% The experiments involved a self-evaluation component (that is, done by competitors on the sequestered and never-published-before test dataset) that incorporated a large number of ISO-compliant live and fake samples. Samples included irises synthesized by modern Generative Adversarial Networks-based models (StyleGAN2 and StyleGAN3) and near-infrared pictures of various artifacts simulating physical iris presentation attacks.

\section{Submitted Algorithms}

\subsection{Team: Fraunhofer IGD}

\paragraph{Method:} 

The Fraunhofer IGD team's submitted algorithm is based on the attention-based pixel-wise binary supervision network (A-PBS) \cite{fang2021iris}, which demonstrated enhanced generalizability of iris PAD on both near-infrared (NIR) and visible spectrum domains \cite{Fang_HoPAD_2023}. A-PBS is comprised of two components: 1) capturing the fine-grained pixel/patch-level cues with the help of pixel-wise binary supervision during training, and 2) automatically localizing the region that contributes the most to an accurate PAD decision by an attention mechanism. The model utilizes two blocks from DenseNet161 and was initialized with the weights trained on the ImageNet dataset. The model was then fine-tuned using iris PAD data with the input size of $224 \times 224$. To increase the diversity of the training data, data augmentation techniques, such as horizontal flipping and Contrast Limited Adaptive Histogram Equalization (CLAHE), were used during training. Two loss functions were combined, where smooth L1 aimed to help the network reduce its sensitivity to outliers in the feature map, and binary cross-entropy loss is used for the binary supervision. The initial learning rate was set to 0.0001 and halved every six epochs with a maximum of 20 epochs. Adam optimizer was used with a weight decay of 0.0005. The final prediction score is the binary output. 

\paragraph{Training data} is composed of five publicly-available iris PAD datasets: CASIA-Iris-Syn \cite{casia-database}, NDCLD'13 \cite{Doyle_BTAS_2013}, NDCLD'15 \cite{Doyle_Access_2015}, and IIITD-WVU and Clarkson subsets of the LivDet-Iris 2017 benchmark \cite{Yambay_IJCB_2017}.

\subsection{Team: Beijing University of Civil Engineering and Architecture, China (BUCEA)}

\paragraph{Method:} BUCEA submitted results for four algorithms. The algorithm that obtained the best results (``BUCEA Algo1'' in Tab. \ref{tab:baseline_table}) is described here. It employs a fusion convolutional neural network, which accepts two inputs: one is the original iris image, and the other is the iris image enhanced by contrast limited adaptive histogram equalization (CLAHE) algorithm. CLAHE is primarily used to highlight iris texture-related features. Both input images are fed into a base network (\eg, ResNet18) to extract their corresponding features and make a binary prediction (bona fide or PAI), respectively. Then, features from the two-stream network are further fused to make a binary prediction. Finally, a score fusion module is employed to combine all middle binary prediction scores to generate a PAD score. 

\paragraph{Training data} is comprised of 74,278 images (43,078 bona fide, 31,200 PAI samples), and combines approximately 10 open-sourced datasets, such as: LivDet-Iris 2017 \cite{Yambay_IJCB_2017}, NDCLD'15 \cite{Doyle_Access_2015}, IIITD Contact Lens Iris \cite{Kohli_ICB_2013}, ND-CrossSensor-Iris-2013, and BUCEA team's self-made print dataset. The dataset contains many types of attacks, such as contact lenses, printouts, synthetic samples, and doll eyes. During training, a weighted sampling was performed to ensure an equal bona fide-to-PAI sample ratio.

\subsection{Team: Hochschule Darmstadt (HDA) and IDVisionCenter}

\paragraph{Method:} Team HDA developed an algorithm based on the Swin Transformer architecture\cite{Swin} with a multi-class linear classifier as the final stage. Four PAI classes were taken into consideration (printed, pattern, dead and synthetic), plus the bona fide class. Input images were transformed according to ImageNet's transforms of RGB images (in this case, the grayscale NIR image was repeated on each channel) and resized to $256 \times 256$ pixels. The model was fine-tuned from ImageNet 1K weights for 50 epochs, and results were computed with a model achieving the best validation results (after 38 epochs). Softmax was used on the classifier's output to get the score of the bona fide class. 

\paragraph{Training data:} The LivDet-Iris 2020 competition database and complement images were used for this method \cite{Tapia,das2020iris}. In addition, three sets of complementary databases of iris images were also utilized. First, a database of NIR bona fide images, captured using an iT TD100 iris sensor with a resolution of $640\times 480$ pixels, called ``Iris-CL1''. A second database, called ``Iris-printed-CL1'', containing high-quality PAI images of printed PAIs, was also created. The database was also complemented with 10,000 synthetic images based on Maureira \etal \cite{Maureira}. Overall, 37,964 images of NIR iris were used for training.

\begin{table}[!ht]
\caption{Numbers and sources of iris images used to train baseline models.}
\tiny
\begin{center}
\label{app:full_dataset}
\begin{tabular}{|c|ccc|}
\toprule
\textbf{Image Type} & \textbf{Contributing Dataset} & \textbf{\# of Samples}& \textbf{Total \#} \\
& & & \textbf{of Samples} \\
\midrule
Bona fide& \begin{tabular}[c]{@{}c@{}}
ATVS-FIr \cite{Galbally_ICB_2012}\\
BERC\_IRIS\_FAKE \cite{Sung_OE_2007}\\
CASIA-Iris-Thousand \cite{casia-database}\\
CASIA-Iris-Twins \cite{casia-database}\\
Disease-Iris v2.1 \cite{Trokielewicz_BTAS_2015}\\
ETPAD v2 {\cite{ETPAD_v2_URL}}\\
IIITD Contact Lens Iris \cite{Kohli_ICB_2013}\\
IIITD Combined Spoofing Database \cite{Kohli_BTAS_2016} \\
LivDet-Iris Clarkson 2015 \cite{Yambay_ISBA_2017} \\
LivDet-Iris Warsaw 2015 \cite{Yambay_ISBA_2017}\\
LivDet-Iris Clarkson 2017 \cite{Yambay_IJCB_2017} \\
LivDet-Iris IIITD-WVU 2017 \cite{Yambay_IJCB_2017}\\
LivDet-Iris Warsaw 2017 \cite{Yambay_IJCB_2017}\\
Notre Dame \\~\end{tabular}
& \begin{tabular}[c]{@{}c@{}}800\\ 2,776\\ 19,952\\ 3,181\\ 255\\ 400\\ 13\\ 4,531\\ 813\\ 36\\ 3,949\\ 2,944\\ 5,167\\ 20,529\\(43,526)$^{**}$ \end{tabular} & \begin{tabular}[c]{@{}c@{}} 70,677\\(93,674)$^{**}$ \end{tabular}\\ \midrule

Artificial & \begin{tabular}[c]{@{}c@{}}BERC\_IRIS\_FAKE \cite{Sung_OE_2007} \\ Notre Dame 
\end{tabular} & \begin{tabular}[c]{@{}c@{}}80\\ 197 \end{tabular} & 277 \\ \midrule

Textured contact lenses  & \begin{tabular}[c]{@{}c@{}}
BERC\_IRIS\_FAKE \cite{Sung_OE_2007}\\
IIITD Contact Lens Iris \cite{Kohli_ICB_2013}\\
LivDet-Iris Clarkson 2015 \cite{Yambay_ISBA_2017} \\
LivDet-Iris Clarkson 2017 \cite{Yambay_IJCB_2017}\\
LivDet-Iris IIITD-WVU 2017 \cite{Yambay_IJCB_2017}\\
Notre Dame \end{tabular}
& \begin{tabular}[c]{@{}c@{}}140\\ 3,420\\ 1,107\\ 1,881\\ 1,700\\ 19,124
\end{tabular}
& 27,372 \\ \midrule

Textured contact & LivDet-Iris IIITD-WVU 2017 \cite{Yambay_IJCB_2017}  & 1,899  & 1,899 \\
lenses \& printed & &  & \\\midrule

Diseased & Disease-Iris v2.1 \cite{Trokielewicz_BTAS_2015} & 1,537 & 1,537 \\ \midrule

Post-mortem & Post-Mortem-Iris v3.0 \cite{Trokielewicz_IVC_2020} & 2,259 & 2,259 \\ \midrule

Paper printouts & \begin{tabular}[c]{@{}c@{}}
ATVS-FIr \cite{Galbally_ICB_2012}\\
BERC\_IRIS\_FAKE \cite{Sung_OE_2007}\\
IIITD Combined Spoofing Database \cite{Kohli_BTAS_2016}\\
LivDet-Iris Clarkson 2015 \cite{Yambay_ISBA_2017}\\
LivDet-Iris Warsaw 2015 \cite{Yambay_ISBA_2017}\\
LivDet-Iris Clarkson 2017 \cite{Yambay_IJCB_2017}\\
LivDet-Iris IIITD-WVU 2017 \cite{Yambay_IJCB_2017}\\
LivDet-Iris Warsaw 2017 \cite{Yambay_IJCB_2017}\end{tabular} & \begin{tabular}[c]{@{}c@{}}800\\ 1,600\\ 1,371\\ 1,745\\ 20\\ 2,250\\ 1,766\\ 6,841\end{tabular} & 16,393 \\ \midrule

Synthetic & CASIA-Iris-Syn V4 \cite{Wei_ICPR_2008} & 10,000 & 10,000  \\ \midrule

GAN Synthetic & StyleGAN2-ADA, StyleGAN3  & 22,997 & 22,997  \\ \midrule

Displayed on e-ink device & --- & 81 & 81 \\

\midrule
{\bf All combined} & & & {\bf 141,352}\\
& & & {\bf (187,348)}$^*$\\
\bottomrule

\end{tabular}
{\footnotesize\vskip1mm
$^{**}$excluding (including) additional bona fide samples to match the number of PAI samples (after adding synthetic images); $^{*}$excluding (including) GAN synthetic images and extra bona fide samples} %(look for: {\it ND-OpenSetIrisPAD-supplemental}; will be added )
\end{center}
\label{tab:dataset-refinement}
\end{table}

\begin{table*}
\centering
\caption{Summary of the LivDet-2023 Iris results. APCER1 represents the APCER as weighted by the number of samples for each PAI type, while APCER2 represents the APCER with equal weights for each PAI type. ACER1 and ACER2 are calculated by averaging BPCER and the respective APCER. The ACER1 and ACER2 of the winners are underlined. Vision Transformer (ViT)-, DenseNet-, Inception- and ResNet-based baseline models were trained in four variants: ``vanilla'' (neither augmentations not synthetic samples were used), ``augs'' (included image augmentations), ``synth'' (included StyleGAN2 and StyleGAN3 samples), and ``both'' (augmentations and GAN synthetic samples added to the training set.}

\footnotesize
\begin{tabular}{|l|c|ccc|ccc|ccc|cc|}
\toprule
& \multicolumn{7}{c}{\bf APCER (\%)} & \multicolumn{3}{c}{\multirow{1}{*}{\bf Overall Metrics (\%) }} & \multicolumn{2}{c|}{\multirow{1}{*}{\bf ACER (\%) }} \\
\multirow{3}{*}{\textbf{Algorithm}} & Level A & \multicolumn{3}{c}{Level B} & \multicolumn{3}{c|}{Level C - Synth} & \multicolumn{3}{c}{  } & \multicolumn{2}{c|}{} \\
& PE & F/P & CL & ED & LQ & MQ & HQ & BPCER & APCER1 & APCER2 & ACER1 & ACER2 \\

\midrule
\multicolumn{13}{c}{{\bf LivDet-Iris 2023 Competing Algorithms}}\\
\midrule
Fraunhofer IGD & 0.57 & 0.00 & 0.00 & 0.00 & 30.00 & 51.35 & 52.50 & 35.40 & 39.22 & 19.20 & {\bf\underline{37.31}} & 27.30 \\
BUCEA Algo1 & 22.80 & 1.50 & 0.00 & 0.00 & 95.55 & 94.40 & 94.85 & 0.14 & 85.16 & 44.16 & 42.65 & {\bf\underline{22.15}} \\
BUCEA Algo3 & 29.12 & 7.14 & 0.00 & 2.50 & 99.20 & 99.10 & 99.35 & 0.18 & 89.64 & 48.06 & 44.91 & 24.12 \\
BUCEA Algo4 & 38.70 & 7.52 & 0.00 & 15.00 & 98.65 & 99.25 & 99.45 & 0.60 & 90.37 & 51.22 & 45.49 & 25.91 \\
BUCEA Algo2 & 51.34 & 25.94 & 0.00 & 27.50 & 97.20 & 98.15 & 97.95 & 0.78 & 90.94 & 56.87 & 45.86 & 28.82 \\
HDA-IDVC Algo1 & 1.53 & 1.13 & 0.00 & 0.00 & 67.15 & 66.00 & 67.80 & 42.34 & 58.98 & 29.09 & 50.66 & 35.71 \\
HDA-IDVC Algo2 & 2.30 & 0.00 & 0.00 & 0.00 & 62.55 & 63.95 & 65.00 & 49.92 & 56.23 & 27.69 & 53.07 & 38.80 \\
HDA-IDVC Algo3 & 2.30 & 0.00 & 0.00 & 0.00 & 62.55 & 63.95 & 65.00 & 49.92 & 56.23 & 27.69 & 53.07 & 38.80 \\
\midrule
\multicolumn{13}{c}{{\bf LivDet-Iris 2023 Baseline Algorithms}}\\
\midrule
DenseNet (vanilla) & 0.00 & 0.00 & 0.00 & 0.00 & 88.45 & 95.85 & 97.25 & 0.34 & 82.41 & 40.22 & 41.37 & 20.28 \\
DenseNet (augs) & 0.00 & 0.38 & 0.00 & 0.00 & 96.50 & 97.60 & 98.85 & 0.17 & 85.76 & 41.90 & 42.97 & 21.04 \\
DenseNet (synths) & 0.00 & 0.00 & 0.00 & 0.00 & 0.10 & 0.50 & 0.15 & 0.46 & 0.22 & 0.11 & 0.34 & 0.28 \\
DenseNet (both) & 0.38 & 0.38 & 0.00 & 0.00 & 0.35 & 0.55 & 0.15 & 0.26 & 0.35 & 0.26 & 0.31 & 0.26\\
\hline\hline
Inception (vanilla) & 0.19 & 1.13 & 0.00 & 0.00 & 96.10 & 98.80 & 99.55 & 0.11 & 86.24 & 42.25 & 43.18 & 21.18 \\ Inception (augs) & 0.00 & 0.38 & 0.00 & 0.00 & 87.00 & 93.65 & 94.35 & 0.31 & 80.51 & 39.34 & 40.41 & 19.83 \\
Inception (synths) & 0.00 & 0.00 & 0.00 & 0.00 & 0.10 & 0.15 & 0.05 & 0.45 & 0.09 & 0.04 & 0.27 & 0.25 \\
Inception (both) & 0.00 & 0.00 & 0.00 & 0.00 & 0.80 & 0.15 & 0.05 & 0.31 & 0.29 & 0.14 & 0.30 & 0.23 \\
\hline\hline
ResNet (vanilla) & 0.00 & 0.00 & 0.00 & 0.00 & 95.85 & 97.45 & 98.00 & 0.15 & 85.26 & 41.61 & 42.71 & 20.88 \\
ResNet (augs) & 1.34 & 0.00 & 0.00 & 2.50 & 97.75 & 98.60 & 99.45 & 0.32 & 86.70 & 42.81 & 43.51 & 21.56 \\
ResNet (synths) & 0.00 & 0.00 & 0.00 & 0.00 & 0.20 & 0.20 & 0.10 & 0.43 & 0.15 & 0.07 & 0.29 & 0.25 \\
ResNet (both) & 0.00 & 0.38 & 0.00 & 0.00 & 0.40 & 1.80 & 1.25 & 0.12 & 1.02 & 0.55 & 0.57 & 0.33 \\
\hline\hline
ViT (vanilla) & 0.00 & 0.00 & 0.00 & 0.00 & 97.35 & 99.00 & 99.55 & 0.15 & 86.61 & 42.27 & 43.38 & 21.21 \\
ViT (augs) & 1.53 & 0.38 & 0.00 & 0.00 & 97.40 & 99.35 & 99.65 & 0.05 & 86.89 & 42.62 & 43.47 & 21.33 \\
ViT (synths) & 0.19 & 0.00 & 0.00 & 0.00 & 2.70 & 1.90 & 0.95 & 0.46 & 1.64 & 0.82 & 1.05 & 0.64 \\
ViT (both) & 0.77 & 0.00 & 0.00 & 0.00 & 11.00 & 7.95 & 4.90 & 0.18 & 7.04 & 3.52 & 3.61 & 1.85 \\
\hline\hline
StyleGAN2 (LQ) & 0.00 & 0.00 & 0.00 & 0.00 & 0.00 & 0.00 & 0.00 & 100.00 & 0.00 & 0.00 & 50.00 & 50.00 \\
StyleGAN2 (MQ) & 0.00 & 0.00 & 0.00 & 0.00 & 0.00 & 0.00 & 0.00 & 100.00 & 0.00 & 0.00 & 50.00 & 50.00 \\
StyleGAN2 (HQ) & 0.00 & 0.00 & 0.00 & 0.00 & 0.00 & 0.00 & 0.00 & 100.00 & 0.00 & 0.00 & 50.00 & 50.00 \\
\hline\hline
StyleGAN3 (LQ) & 35.25 & 23.68 & 25.00 & 32.50 & 6.00 & 8.05 & 7.75 & 96.17 & 10.20 & 19.75 & 53.19 & 57.96 \\
StyleGAN3 (MQ) & 0.00 & 0.00 & 0.00 & 0.00 & 0.00 & 0.00 & 0.00 & 100.00 & 0.00 & 0.00 & 50.00 & 50.00 \\
StyleGAN3 (HQ) & 0.00 & 0.00 & 0.00 & 0.00 & 0.00 & 0.00 & 0.00 & 100.00 & 0.00 & 0.00 & 50.00 & 50.00 \\
\hline\hline
Xception (Xent) & 11.69 & 62.41 & 0.00 & 7.50 & 87.15 & 95.90 & 96.10 & 2.63 & 85.07 & 51.54 & 43.85 & 27.08 \\
Xception (CYBORG) & 18.20 & 40.98 & 25.00 & 17.50 & 77.55 & 87.40 & 88.30 & 7.79 & 77.23 & 50.70 & 42.51 & 29.25 \\

\midrule
\multicolumn{13}{c}{{\bf LivDet-Iris 2023 Human Examination}}\\
\midrule
Human subjects & 30.12 & 7.06 & 20.41 & 17.57 & 33.19 & 50.25 & 51.14 & 38.38 & 40.98 & 29.96 & 39.68 & 34.17 \\
\bottomrule
\end{tabular}
\label{tab:baseline_table}
\end{table*}

\section{Baseline Algorithms}

\subsection{Architectures}

\paragraph{ResNet \cite{he2015deep}} is known for its residual connections, which allow the network to learn residual representations, help in alleviating the vanishing gradient problem and enable training deeper networks. By learning the residuals, the network can focus on fine-tuning the learned features instead of re-learning them from scratch, facilitating the training. ResNet101 was used in this work. 

\paragraph{DenseNet \cite{huang2017densely}} is a CNN-based model, in which each layer receives direct input from all preceding layers within a dense block, encouraging feature reuse and flow of gradients throughout the network. In this work, a pre-trained DenseNet121 was used. 

\paragraph{Inception \cite{szegedy2017inception}} utilizes multiple parallel convolutional operations of different kernel sizes to capture features at different scales and thus facilitate learning both local and global patterns. In this work, a pre-trained Inception\_v3 network was used, which incorporates auxiliary classifiers during training for intermediate supervision mitigating the vanishing gradient problem through additional gradient flow and regularization. 

\paragraph{Vision Transformer (ViT) \cite{dosovitskiy2020image}} applies the transformer model, successful in natural language processing, to the computer vision domain. In ViT, an input image is divided into fixed-size patches, which are then processed by transformer layers. In this work, we used the pre-trained ViT\_b\_16, which is a smaller-sized model with a patch size of 16. This model consists of a stack of transformer encoder layers, each having a multi-head self-attention mechanism and position-wise feed-forward network.  
In the ViT, the whole image is considered a single token and serves as the entry point for the transformer. It also uses a classification head on top of the transformer stack to make the PAD-related predictions. One of the major advantages of ViT architectures is their ability to capture long-range dependencies and context information. 

\paragraph{Xception \cite{chollet2017xception}} relies on depth-wise separable convolutions and uses the same skip connections as seen in the ResNet architecture. The application of depth-wise and point-wise convolutions helps reduce computational complexity as well as training time while increasing classification accuracy. We built two Xception-based baselines. Once trained traditionally (with cross-entropy loss). The second way of training followed the approach of Boyd \etal's CYBORG paradigm~\cite{boyd2021cyborg}, which incorporates human perceptual expertise into the training process. In Boyd \etal's work, to collect human perception data used also in our training, Mechanical Turk workers were shown bona fide and PAI iris images and asked to highlight regions of the image that supported their classification decision. During training, these human annotations were compared to the model's class activation mapping, penalizing the differences between human's and model's saliency maps. By doing so, the models were encouraged to focus on image regions judged as salient by humans.

\paragraph{StyleGAN discriminators} %Although typically used and publicized for their image-generating abilities, GANs 
also offer a discriminator model whose main function is to discern ``fake'' from ``bona fide'' samples. For this competition, six different generative models were used for iris image synthesis: StyleGAN2-LQ, StyleGAN2-MQ, StyleGAN2-HQ, StyleGAN3-LQ, StyleGAN3-MQ, StyleGAN3-HQ (LQ = low quality, MQ = medium quality, and HQ = high quality). The quality of these models was assigned based on the FID score during the training process as well as a qualitative visual inspection of the corresponding image snapshots. The GAN-sourced discriminator models produce logits and thus can be evaluated in the same manner as the competitors' submissions and other baseline models. 
\begin{figure*}[!ht]
\centering

\subcaptionbox{\centering All PAI Categories}{\includegraphics[width=0.245\textwidth]{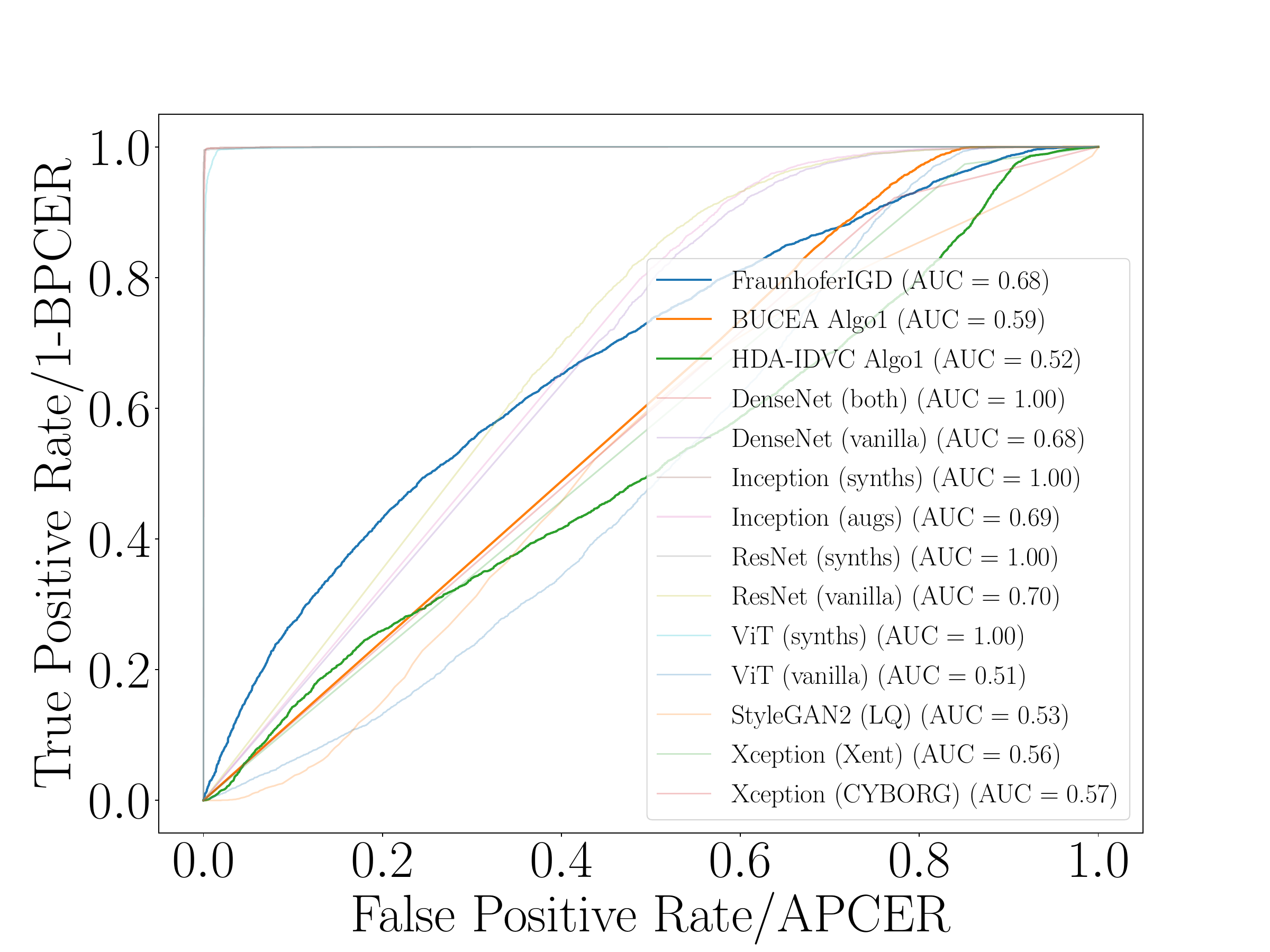}}
\hfill
\subcaptionbox{\centering Printed Eyes (PE)}{\includegraphics[width=0.245\textwidth]{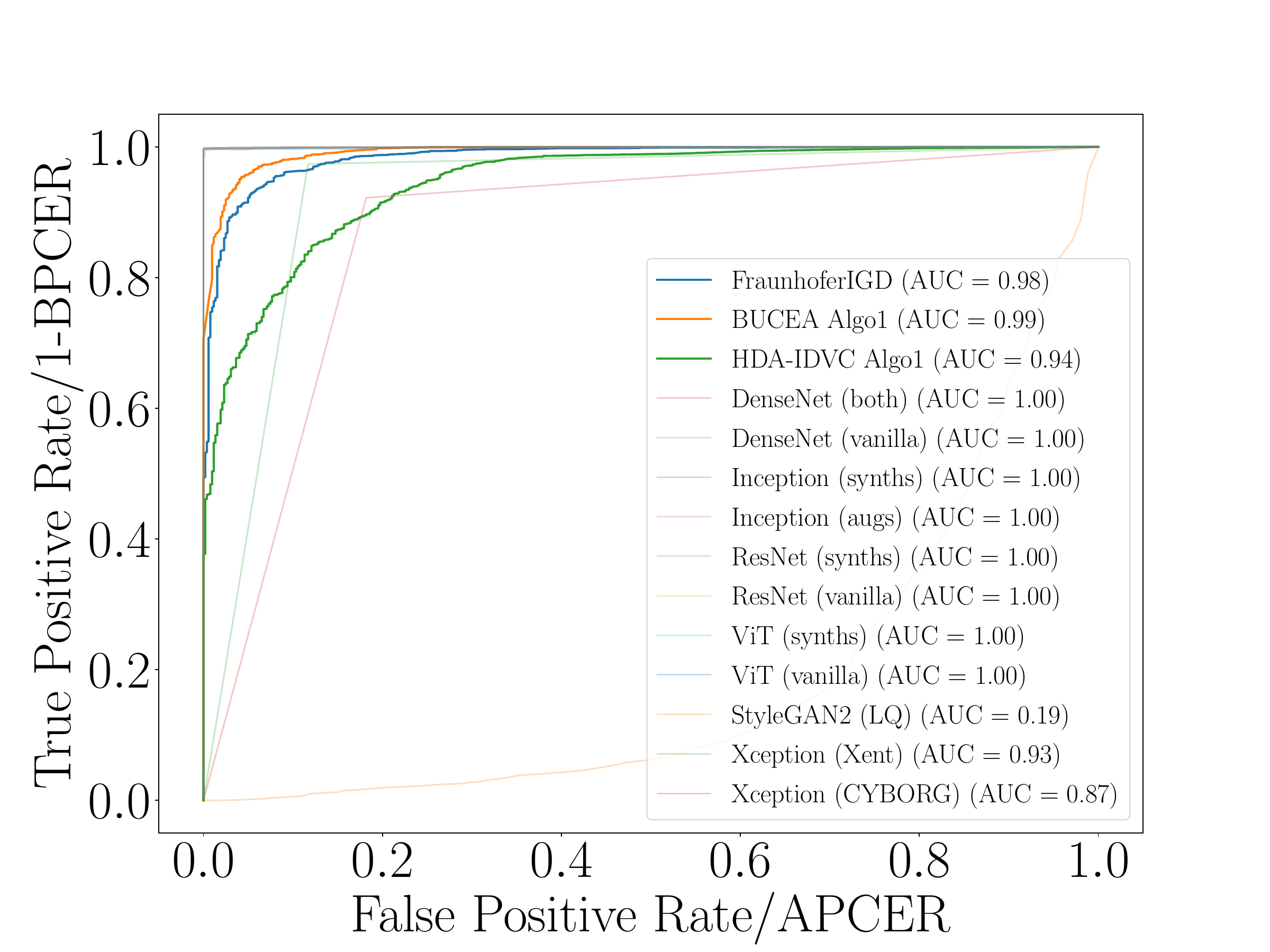}}
\hfill
\subcaptionbox{\centering Fake/Prosthetic Eyes with Add-Ons (FP)}{\includegraphics[width=0.245\textwidth]{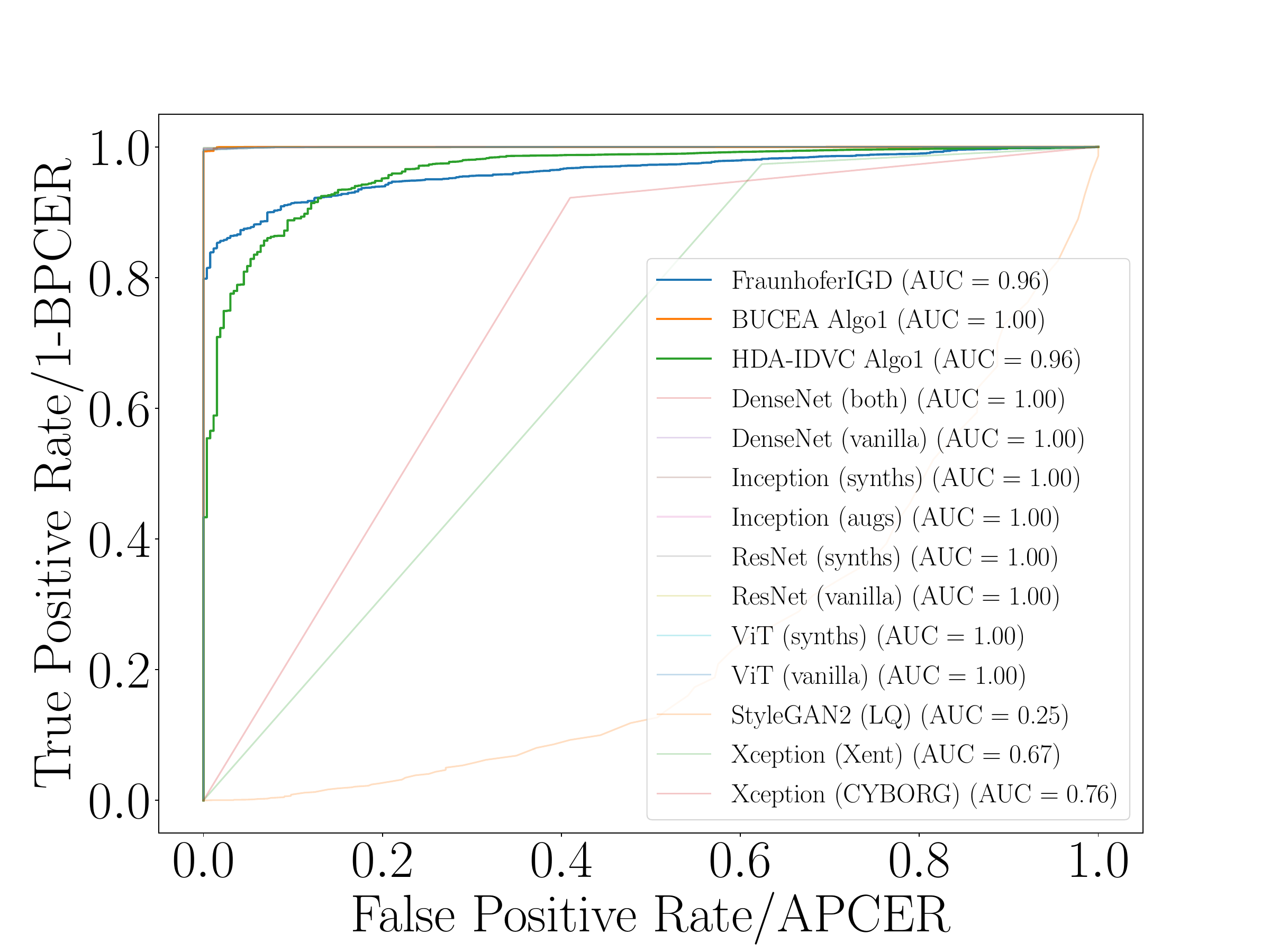}}
\hfill
\subcaptionbox{\centering Electronic Display (ED)}{\includegraphics[width=0.245\textwidth]{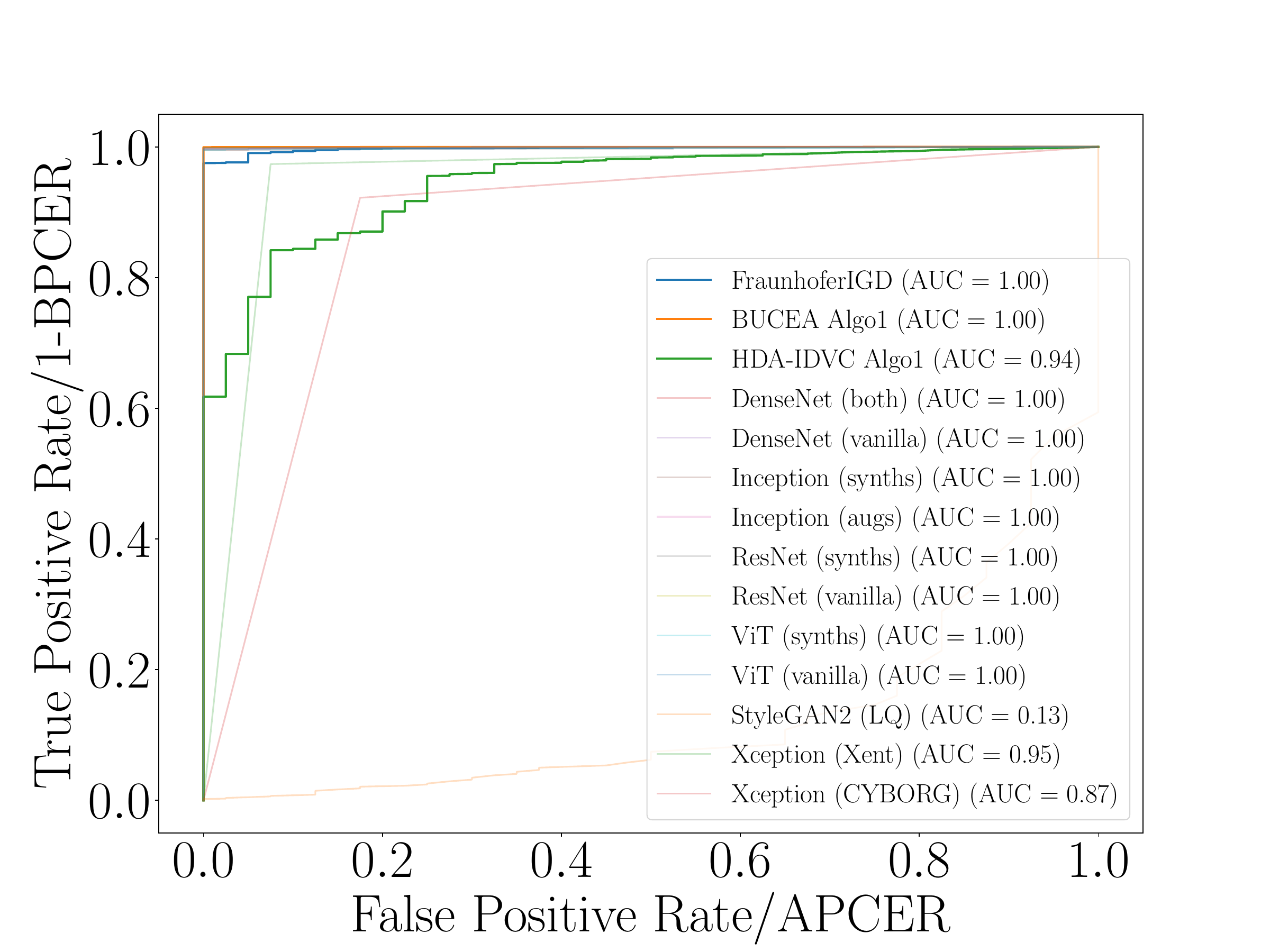}}
\hfill
\newline
\subcaptionbox{\centering Synthetic Iris -- Low Quality (LQ)}{\includegraphics[width=0.24\textwidth]{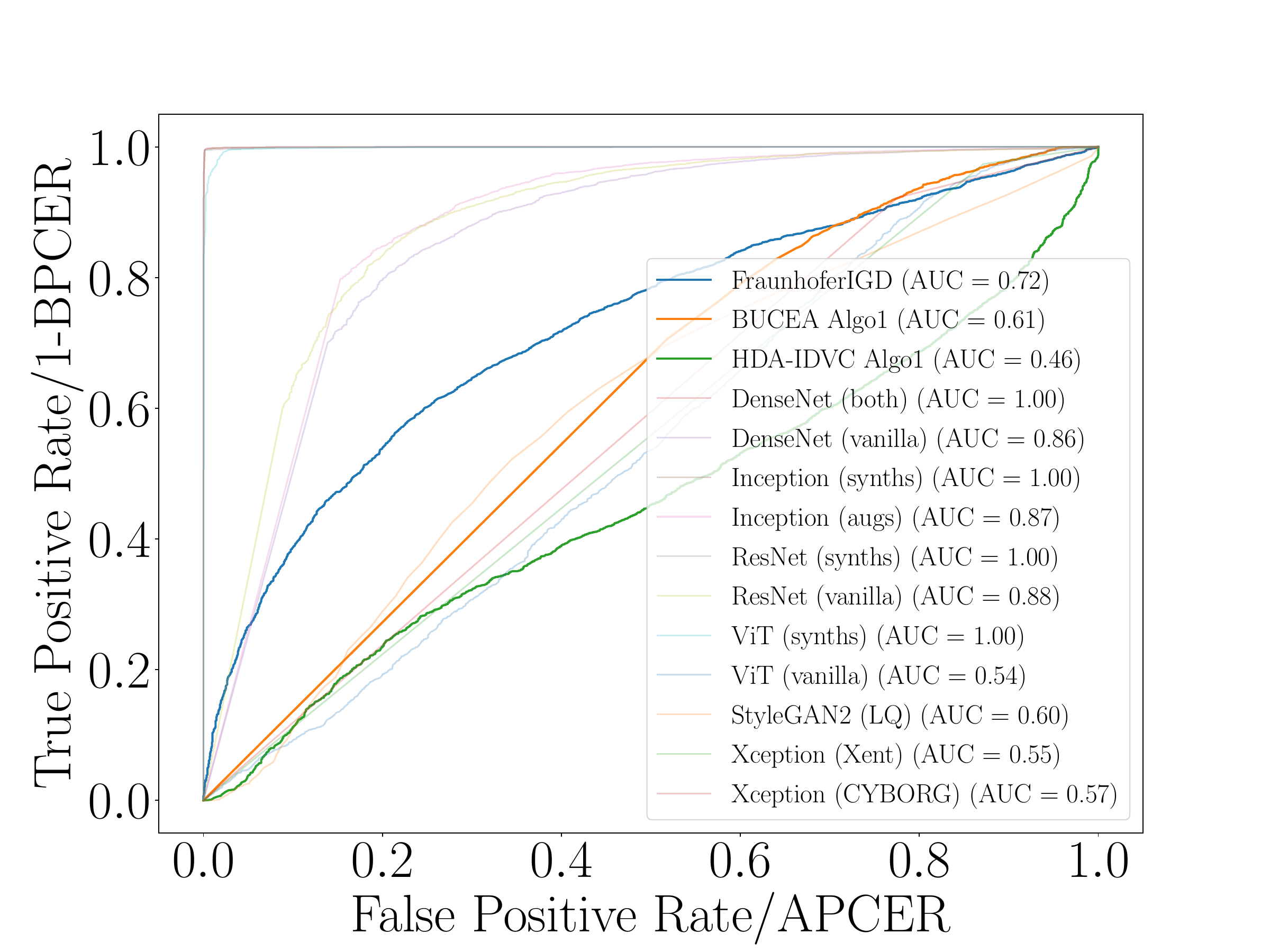}}
\subcaptionbox{\centering Synthetic Iris -- Medium Quality (MQ)}{\hskip6mm\includegraphics[width=0.24\textwidth]{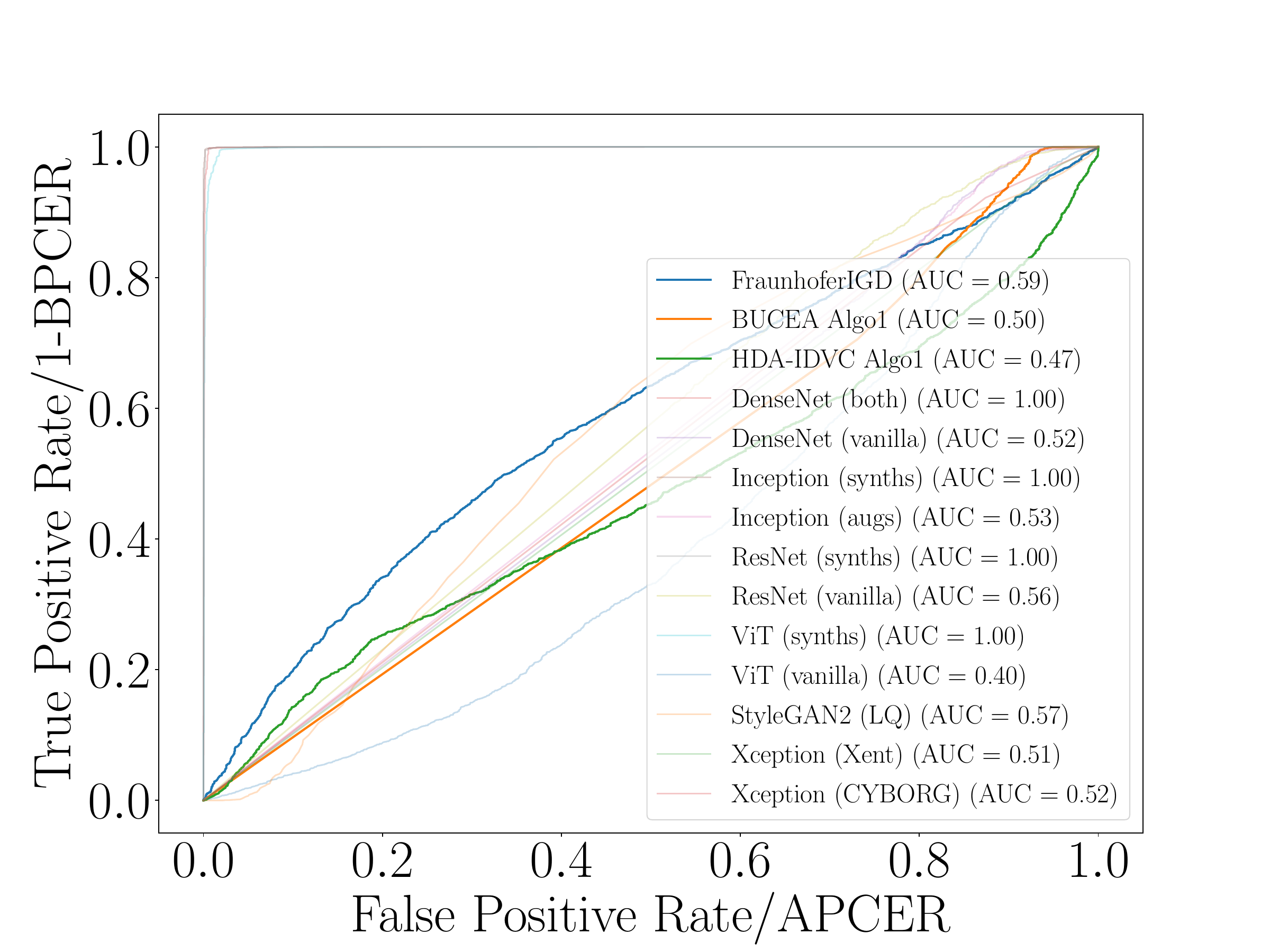}\hskip6mm}
\subcaptionbox{\centering Synthetic Iris -- High Quality (HQ)}{\hskip6mm\includegraphics[width=0.24\textwidth]{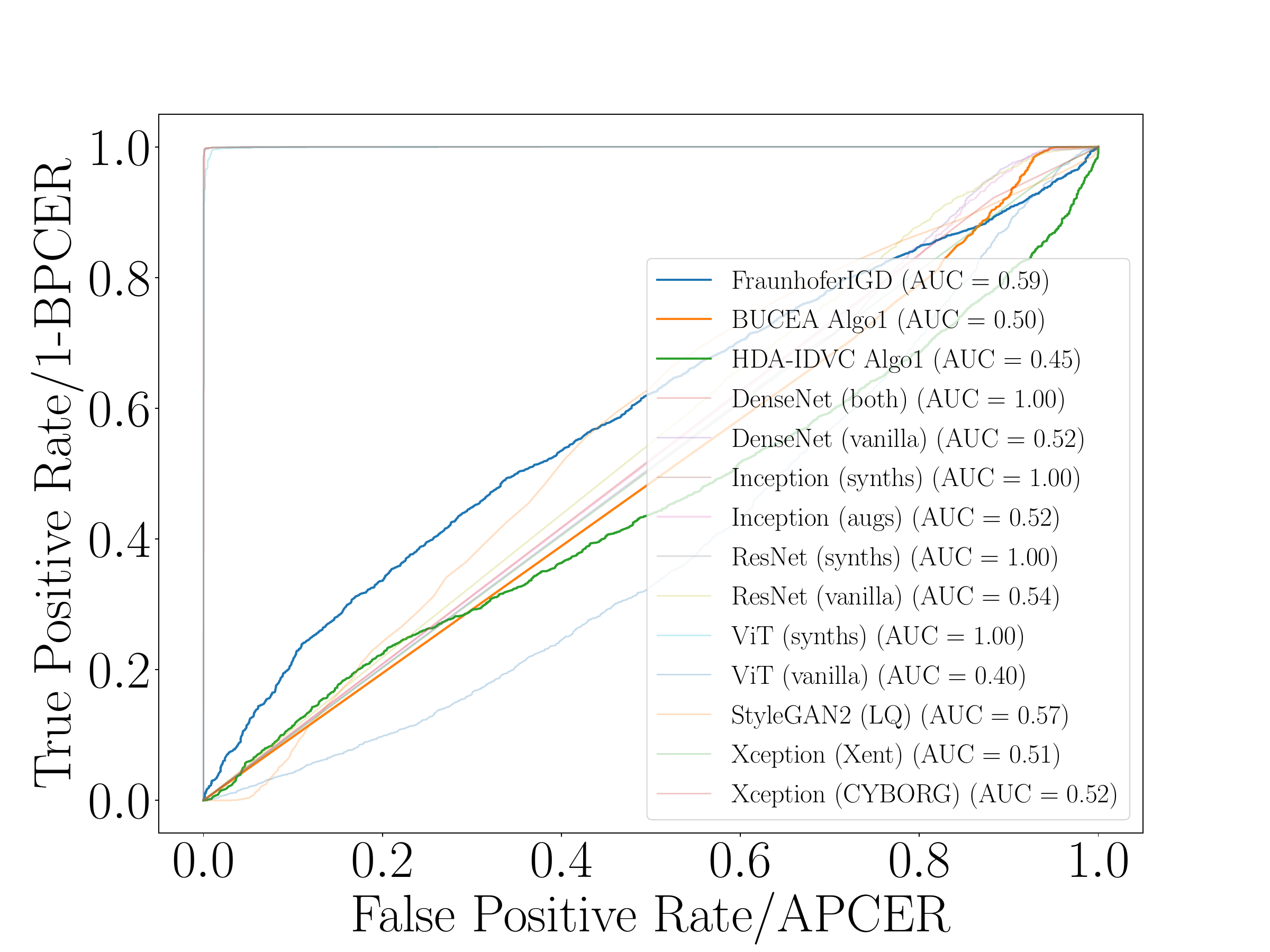}\hskip6mm}
\vskip2mm
\caption{ROC curves for top-performing algorithms from each team as well as trained baseline models. ROCs are shown for all samples (top-left) and each PAI (as labeled). The overall ACER1 is evaluated based on sample-weighted APCER (APCER1). The operating point (``O.P. at 0.5'') was used to rank participants of this LivDet-Iris competition.}
\label{fig:ROC_ALL}
\end{figure*}

\subsection{Training Baseline Models}

Four of the baseline models (ResNet, DenseNet, Inception, and ViT) were trained under two different scenarios: with and without GAN-generated samples. The purpose of this experiment was to see whether the inclusion of samples generated by the same source (thereby making the problem a closed-set problem, from the PAI category point of view) changes the PAD performance significantly. For each scenario, baseline models were trained with and without data augmentation; hence, for each baseline model, we had 4 different training runs: 1) without augmentation and without GAN synthetic samples (``vanilla'' in Tab. \ref{tab:baseline_table}), 2) with augmentation and without GAN synthetic samples (``augs'' in Tab. \ref{tab:baseline_table}), 3) without augmentation and with GAN synthetic samples (``synths'' in Tab. \ref{tab:baseline_table}), and 4) with augmentation and with GAN synthetics (``both'' in Tab. \ref{tab:baseline_table}). The augmentation techniques were adopted from \cite{imgaug} and included affine transformations, randomly selected horizontal flipping, sharpening, blurring, and adding Laplace noise, contrast, and brightness. The training data was made up of iris images (bona fide and PAs) from the largest corpus available. Without synthetic samples, training data contained 70,677 bona fide samples and 70,677 PAI samples; with StyleGAN samples, the data contained 93,674 bona fide images and the same number for PAI samples. Table \ref{tab:dataset-refinement} presents the details of the datasets that we used for training the baseline.  We used 80\% of the entire datasets for training and 20\% for testing the model during training. Each model was trained for 50 epochs, the batch size was set to 32, Cross-entropy and SGD were chosen as the loss and optimizer for all of the models. All of the baseline models were trained on cropped iris images. 

\subsection{Human Examination}

A human examination of the 13,332 test samples was conducted via Prolific's online tool \cite{prolific}. In total, 300 subjects were asked to annotate 50 images each, covering the entire set of test images. Subjects were shown an image (either bona fide or PA) and asked to select the nature of the supplied image in a two-alternative forced choice (2AFC) manner. Results can be seen in Table \ref{tab:baseline_table}.

\section{Results}

\paragraph{Competition Winners:}
%This year's edition of the LivDet-Iris competition
LivDet-Iris 2023 has two winners, corresponding to the two different definitions of ACER (see Table \ref{tab:baseline_table}). According to the first definition (ACER1), which weights APCERs by the number of samples per PAI category, the Fraunhofer IGD team wins, with an ACER1 of 37.31\%. For the second definition of ACER (ACER2), which weights all PAI categories equally, the BUCEA team wins with Algo 1, with an ACER2 of 22.15\%. 
% The BUCEA team's winning algorithm showed a remarkably low BPCER of 0.14\%, suggesting that live images were rarely classified as spoof.

\paragraph{Closed Set versus Open Set:} As seen in Figure \ref{fig:ROC_ALL}, all baseline models trained in a closed set fashion outperformed their open set counterparts. The models that saw synthetic samples during training (labeled either ``both'' or ``synths'') showed ACER values under $<$2.0\% (see Tab. \ref{tab:baseline_table}) and Area Under the ROC Curve (AUC) above 99.9\%. Since this discrepancy in performance is consistent across all network architectures, it seems that StyleGAN-generated samples are still detectable, but only in the less-than-realistic closed-set experimental setup. This result is corroborated in iris PAD literature \cite{BoydIrisPAD}. 

\paragraph{Physical Spoofs versus Synthetic Spoofs:}
For all submitted algorithms, synthetic samples were harder to detect than  physical samples. For example, team Fraunhofer IGD's submitted algorithm showed an APCER1 of 0.044\% on physical attacks and an APCER1 of 39.177\% on synthetic attacks. Unlike physical samples, GAN-generated samples are entirely digital, meaning they pose the largest threat when injected directly into machine learning models.

\paragraph{Accuracy of Human Examiners in iris PAD:} Human examiners (who likely have not seen GAN-generated irises before) performed poorly on the StyleGAN samples, as seen in Table \ref{tab:baseline_table}. In the case of the LQ samples, artifacts such as bubbly textures, misshapen pupils, and smudged eyelashes served as clear signals to humans that an image was a PAI, as evidenced by the relatively lower APCER of \~33\%. However, MQ and HQ samples yielded APCER values closer to that of a random chance classifier (50\%), signifying that images from longer-trained generator models are realistic enough to avoid detection by humans. On the other hand, images of prosthetic or fake irises were the most easily classified by the
%Prolific workers.
human annotators. The results from the
% Prolific
human annotation experiments showed an APCER1 of 39.68\%, which was only beaten by the competition's best algorithm from the Fraunhofer IGD team. 

\paragraph{Accuracy of Human-Trained Models in iris PAD:} In the case of the Xception backbone, the CYBORG-trained model slightly increased performance when compared to the model trained with a cross-entropy loss. So, while humans (in general) struggled with the task of iris PAD, the annotations collected by Boyd \etal in \cite{boyd2022human} ultimately delivered a model with a lower ACER compared to the traditionally-trained model.

\section{Conclusions}
PAD is needed to deter attacks on biometric sensors. This edition of the LivDet-Iris competition has eight competitor algorithms. PAI species included textured contact lenses, PAI electronic display, fake/prosthetic/printed eyes with add-ons as well as synthetic irises of low, medium, and high quality. The winners had an average classification error of 37.31\%, with APCER of 39.22\% and BPCER of 35.40\% with APCER weighted by the number of PAI samples within a species, and an average classification error of 22.15\%, with APCER of 44.16\% and BPCER of 0.14\% with APCER equally weighted across PAI species. Results from multiple baseline algorithms were included with those trained in a closed set regimen obtaining excellent results (ACER $<0.02$), while those trained without all PAI seen during training performed on par with the competition algorithms. Additionally, human examiners performed with an ACER of 39.68\%. This competition represents of state of the art for iris PAD, as well as provides the dataset for further benchmarking after the competition.

\paragraph{Acknowledgements:} This material is based upon work supported by the National Science Foundation under grants No. 2237880 and 1650503. Any opinions, findings, conclusions, or recommendations expressed in this material are those of the author(s) and do not necessarily reflect the views of the National Science Foundation.

{\small
\bibliographystyle{ieee}
\bibliography{bibliography}

\begin{thebibliography}{10}\itemsep=-1pt

\bibitem{prolific}
Prolific - quickly find research participants you can trust.
\newblock \url{http://https://www.prolific.co/}.

\bibitem{boyd2022human}
A.~Boyd, K.~W. Bowyer, and A.~Czajka.
\newblock Human-aided saliency maps improve generalization of deep learning.
\newblock In {\em Proceedings of the IEEE/CVF Winter Conference on Applications of Computer Vision}, pages 2735--2744, 2022.

\bibitem{boyd2020iris}
A.~Boyd, Z.~Fang, A.~Czajka, and K.~W. Bowyer.
\newblock Iris presentation attack detection: Where are we now?
\newblock {\em Pattern Recognition Letters}, 138:483--489, 2020.

\bibitem{BoydIrisPAD}
A.~Boyd, J.~Speth, L.~Parzianello, K.~W. Bowyer, and A.~Czajka.
\newblock Comprehensive study in open-set iris presentation attack detection.
\newblock {\em IEEE Transactions on Information Forensics and Security}, 18:3238--3250, 2023.

\bibitem{boyd2021cyborg}
A.~Boyd, P.~Tinsley, K.~Bowyer, and A.~Czajka.
\newblock Cyborg: Blending human saliency into the loss improves deep learning.
\newblock {\em arXiv preprint arXiv:2112.00686}, 2021.

\bibitem{chollet2017xception}
F.~Chollet.
\newblock Xception: Deep learning with depthwise separable convolutions.
\newblock In {\em Proceedings of the IEEE conference on computer vision and pattern recognition}, pages 1251--1258, 2017.

\bibitem{czajka2018presentation}
A.~Czajka and K.~W. Bowyer.
\newblock Presentation attack detection for iris recognition: An assessment of the state-of-the-art.
\newblock {\em ACM Computing Surveys (CSUR)}, 51(4):1--35, 2018.

\bibitem{das2020iris}
P.~Das, J.~McFiratht, Z.~Fang, A.~Boyd, G.~Jang, A.~Mohammadi, S.~Purnapatra, D.~Yambay, S.~Marcel, M.~Trokielewicz, et~al.
\newblock Iris liveness detection competition (livdet-iris)-the 2020 edition.
\newblock In {\em 2020 IEEE International Joint Conference on Biometrics (IJCB)}, pages 1--9. IEEE, 2020.

\bibitem{dosovitskiy2020image}
A.~Dosovitskiy, L.~Beyer, A.~Kolesnikov, D.~Weissenborn, X.~Zhai, T.~Unterthiner, M.~Dehghani, M.~Minderer, G.~Heigold, S.~Gelly, et~al.
\newblock An image is worth 16x16 words: Transformers for image recognition at scale.
\newblock {\em arXiv preprint arXiv:2010.11929}, 2020.

\bibitem{Doyle_Access_2015}
J.~S. Doyle and K.~W. Bowyer.
\newblock Robust detection of textured contact lenses in iris recognition using bsif.
\newblock {\em IEEE Access}, 3:1672--1683, 2015.

\bibitem{Doyle_BTAS_2013}
J.~S. Doyle, K.~W. Bowyer, and P.~J. Flynn.
\newblock Variation in accuracy of textured contact lens detection based on sensor and lens pattern.
\newblock In {\em 2013 IEEE Sixth International Conference on Biometrics: Theory, Applications and Systems (BTAS)}, pages 1--7, 2013.

\bibitem{fang2021iris}
M.~Fang, N.~Damer, F.~Boutros, F.~Kirchbuchner, and A.~Kuijper.
\newblock Iris presentation attack detection by attention-based and deep pixel-wise binary supervision network.
\newblock In {\em 2021 IEEE International Joint Conference on Biometrics (IJCB)}, pages 1--8. IEEE, 2021.

\bibitem{Galbally_ICB_2012}
J.~Galbally, J.~Ortiz-Lopez, J.~Fierrez, and J.~Ortega-Garcia.
\newblock Iris liveness detection based on quality related features.
\newblock In {\em 2012 5th IAPR Int. Conf. on Biometrics (ICB)}, pages 271--276, New Delhi, India, March 2012. IEEE.

\bibitem{he2015deep}
K.~He, X.~Zhang, S.~Ren, and J.~Sun.
\newblock Deep residual learning.
\newblock {\em Image Recognition}, 7, 2015.

\bibitem{huang2017densely}
G.~Huang, Z.~Liu, L.~Van Der~Maaten, and K.~Q. Weinberger.
\newblock Densely connected convolutional networks.
\newblock In {\em Proceedings of the IEEE conference on computer vision and pattern recognition}, pages 4700--4708, 2017.

\bibitem{casia-database}
{Institute of Automation, Chinese Academy of Sciences}.
\newblock {CASIA Biometrics Database Collections}.
\newblock http://biometrics.idealtest.org.
\newblock Accessed: June 09, 2023.

\bibitem{ISO_IEC_301073:2017}
{ISO/IEC 30107-3}.
\newblock {Information technology -- Biometric presentation attack detection -- Part 3: Testing and reporting}, 2016.

\bibitem{imgaug}
A.~B. Jung, K.~Wada, J.~Crall, S.~Tanaka, J.~Graving, C.~Reinders, S.~Yadav, J.~Banerjee, G.~Vecsei, A.~Kraft, Z.~Rui, J.~Borovec, C.~Vallentin, S.~Zhydenko, K.~Pfeiffer, B.~Cook, I.~Fernández, F.-M. De~Rainville, C.-H. Weng, A.~Ayala-Acevedo, R.~Meudec, M.~Laporte, et~al.
\newblock {imgaug}.
\newblock \url{https://github.com/aleju/imgaug}, 2020.
\newblock Online; accessed 01-Feb-2020.

\bibitem{karras2020sg2ada}
T.~Karras, M.~Aittala, J.~Hellsten, S.~Laine, J.~Lehtinen, and T.~Aila.
\newblock Training generative adversarial networks with limited data.
\newblock In {\em Training Generative Adversarial Networks with Limited Data}, 2020.

\bibitem{karras2021sg3}
T.~Karras, M.~Aittala, S.~Laine, E.~H\"ark\"onen, J.~Hellsten, J.~Lehtinen, and T.~Aila.
\newblock Alias-free generative adversarial networks.
\newblock {\em Proc. NeurIPS}, 2021.

\bibitem{Kohli_ICB_2013}
N.~Kohli, D.~Yadav, M.~Vatsa, and R.~Singh.
\newblock Revisiting iris recognition with color cosmetic contact lenses.
\newblock In {\em {IEEE} Int. Conf. on Biometrics (ICB)}, pages 1--7, Madrid, Spain, June 2013. IEEE.

\bibitem{Kohli_BTAS_2016}
N.~Kohli, D.~Yadav, M.~Vatsa, R.~Singh, and A.~Noore.
\newblock Detecting medley of iris spoofing attacks using desist.
\newblock In {\em {IEEE} Int. Conf. on Biometrics: Theory Applications and Systems (BTAS)}, pages 1--6, Niagara Falls, NY, USA, Sept 2016. IEEE.

\bibitem{ETPAD_v2_URL}
O.~Komogortsev.
\newblock Eye tracker print-attack database (etpad) v2.
\newblock In {\em Eye Tracker Print-Attack Database (ETPAD) v2}, 2014.

\bibitem{Sung_OE_2007}
S.~J. Lee, K.~R. Park, Y.~J. Lee, K.~Bae, and J.~H. Kim.
\newblock {Multifeature-based fake iris detection method}.
\newblock {\em Optical Engineering}, 46(12):1 -- 10, 2007.

\bibitem{Swin}
Z.~Liu, Y.~Lin, Y.~Cao, H.~Hu, Y.~Wei, Z.~Zhang, S.~Lin, and B.~Guo.
\newblock Swin transformer: Hierarchical vision transformer using shifted windows.
\newblock {\em arXiv preprint arXiv:2103.14030}, 2021.

\bibitem{Marcialis_ICIAP_2009}
G.~L. Marcialis, A.~Lewicke, B.~Tan, P.~Coli, D.~Grimberg, A.~Congiu, A.~Tidu, F.~Roli, and S.~Schuckers.
\newblock First international fingerprint liveness detection competition---livdet 2009.
\newblock In P.~Foggia, C.~Sansone, and M.~Vento, editors, {\em Image Analysis and Processing -- ICIAP 2009}, pages 12--23, Berlin, Heidelberg, 2009. Springer Berlin Heidelberg.

\bibitem{Maureira}
J.~Maureira, J.~E. Tapia, C.~Arellano, and C.~Busch.
\newblock Analysis of the synthetic periocular iris images for robust presentation attacks detection algorithms.
\newblock {\em IET Biometrics}, 11(4):343--354, 2022.

\bibitem{Fang_HoPAD_2023}
N.~D. Meiling~Fang, Fadi~Boutros.
\newblock {\em Handbook of Biometric Anti-Spoofing: Presentation Attack Detection and Vulnerability Assessment}, chapter Intra and Cros-spectrum Iris Presentation Attack Detection in the NIR and Visible Domains, pages 171--199.
\newblock Springer, third edition, 2023.

\bibitem{szegedy2017inception}
C.~Szegedy, S.~Ioffe, V.~Vanhoucke, and A.~Alemi.
\newblock Inception-v4, inception-resnet and the impact of residual connections on learning.
\newblock In {\em Proceedings of the AAAI conference on artificial intelligence}, volume~31, 2017.

\bibitem{Tapia}
J.~E. Tapia, S.~Gonzalez, and C.~Busch.
\newblock Iris liveness detection using a cascade of dedicated deep learning networks.
\newblock {\em IEEE Transactions on Information Forensics and Security}, 17:42--52, 2022.

\bibitem{Trokielewicz_BTAS_2015}
M.~{Trokielewicz}, A.~{Czajka}, and P.~{Maciejewicz}.
\newblock Assessment of iris recognition reliability for eyes affected by ocular pathologies.
\newblock In {\em {IEEE} Int. Conf. on Biometrics: Theory Applications and Systems (BTAS)}, pages 1--6, 2015.

\bibitem{Trokielewicz_IVC_2020}
M.~Trokielewicz, A.~Czajka, and P.~Maciejewicz.
\newblock Post-mortem iris recognition with deep-learning-based image segmentation.
\newblock {\em Image and Vision Computing}, 94:103866, 2020.

\bibitem{Wei_ICPR_2008}
Z.~Wei, T.~Tan, and Z.~Sun.
\newblock Synthesis of large realistic iris databases using patch-based sampling.
\newblock In {\em Int. Conf. on Pattern Recognition (ICPR)}, pages 1--4, Tampa, FL, USA, Dec 2008. IEEE.

\bibitem{Yambay_IJCB_2017}
D.~Yambay, B.~Becker, N.~Kohli, D.~Yadav, A.~Czajka, K.~W. Bowyer, S.~Schuckers, R.~Singh, M.~Vatsa, A.~Noore, D.~Gragnaniello, C.~Sansone, L.~Verdoliva, L.~He, Y.~Ru, H.~Li, N.~Liu, Z.~Sun, and T.~Tan.
\newblock {LivDet Iris 2017} -- iris liveness detection competition 2017.
\newblock In {\em {IEEE} Int. Joint Conf. on Biometrics (IJCB)}, pages 1--6, Denver, CO, USA, 2017. IEEE.

\bibitem{yambay2017livdet}
D.~{Yambay}, B.~{Becker}, N.~{Kohli}, D.~{Yadav}, A.~{Czajka}, K.~W. {Bowyer}, S.~{Schuckers}, R.~{Singh}, M.~{Vatsa}, A.~{Noore}, D.~{Gragnaniello}, C.~{Sansone}, L.~{Verdoliva}, L.~{He}, Y.~{Ru}, H.~{Li}, N.~{Liu}, Z.~{Sun}, and T.~{Tan}.
\newblock {LivDet-Iris 2017 -- Iris Liveness Detection Competition 2017}.
\newblock In {\em IEEE International Joint Conference on Biometrics (IJCB)}, pages 733--741, 2017.

\bibitem{yambay2014schuckers}
D.~{Yambay}, J.~S. {Doyle}, K.~W. {Bowyer}, A.~{Czajka}, and S.~{Schuckers}.
\newblock {LivDet-Iris 2013 -- Iris Liveness Detection Competition 2013}.
\newblock In {\em IEEE International Joint Conference on Biometrics}, pages 1--8, 2014.

\bibitem{Yambay_BTAS_2013}
D.~Yambay, J.~S. Doyle, K.~W. Bowyer, A.~Czajka, and S.~Schuckers.
\newblock Livdet-iris 2013 - iris liveness detection competition 2013.
\newblock In {\em IEEE International Joint Conference on Biometrics}, pages 1--8, 2014.

\bibitem{7947701}
D.~{Yambay}, B.~{Walczak}, S.~{Schuckers}, and A.~{Czajka}.
\newblock {LivDet-Iris 2015 -- Iris Liveness Detection Competition 2015}.
\newblock In {\em IEEE International Conference on Identity, Security and Behavior Analysis (ISBA)}, pages 1--6, 2017.

\bibitem{Yambay_ISBA_2017}
D.~Yambay, B.~Walczak, S.~Schuckers, and A.~Czajka.
\newblock Livdet-iris 2015 - iris liveness detection competition 2015.
\newblock In {\em {IEEE} Int. Conf. on Identity, Security and Behavior Analysis (ISBA)}, pages 1--6, New Delhi, India, Feb 2017. IEEE.

\end{thebibliography}
}

\end{document}